\documentclass[10pt,twocolumn,letterpaper]{article}

\usepackage[pagenumbers]{iccv} 

\usepackage[accsupp]{axessibility}

\usepackage{multirow}
\usepackage{booktabs}
\usepackage{amssymb}
\usepackage{makecell}
\usepackage{float}
\usepackage{makecell} 

\definecolor{iccvblue}{rgb}{0.21,0.49,0.74}
\usepackage[pagebackref,breaklinks,colorlinks,allcolors=iccvblue]{hyperref}

\def\paperID{23} 
\def\confName{ICCV}
\def\confYear{2025}

\title{GasTwinFormer: A Hybrid Vision Transformer for Livestock Methane Emission Segmentation and Dietary Classification in Optical Gas Imaging}

\author{
Toqi Tahamid Sarker\textsuperscript{1},
Mohamed Embaby\textsuperscript{2},
Taminul Islam\textsuperscript{1},
Amer AbuGhazaleh\textsuperscript{1},
Khaled R Ahmed\textsuperscript{1}\\
\textsuperscript{1}Southern Illinois University Carbondale, USA \quad
\textsuperscript{2}University of California, Davis, USA\\
{\tt\small \{toqitahamid.sarker, taminul.islam, aabugha, khaled.ahmed\}@siu.edu, membaby@ucdavis.edu}
}

\begin{document}
\maketitle

\begin{abstract}
Livestock methane emissions represent 32\% of human-caused methane production, making automated monitoring critical for climate mitigation strategies. We introduce GasTwinFormer, a hybrid vision transformer for real-time methane emission segmentation and dietary classification in optical gas imaging through a novel Mix Twin encoder alternating between spatially-reduced global attention and locally-grouped attention mechanisms. Our architecture incorporates a lightweight LR-ASPP decoder for multi-scale feature aggregation and enables simultaneous methane segmentation and dietary classification in a unified framework. We contribute the first comprehensive beef cattle methane emission dataset using OGI, containing 11,694 annotated frames across three dietary treatments. GasTwinFormer achieves 74.47\% mIoU and 83.63\% mF1 for segmentation while maintaining exceptional efficiency with only 3.348M parameters, 3.428G FLOPs, and 114.9 FPS inference speed. Additionally, our method achieves perfect dietary classification accuracy (100\%), demonstrating the effectiveness of leveraging diet-emission correlations. Extensive ablation studies validate each architectural component, establishing GasTwinFormer as a practical solution for real-time livestock emission monitoring. Please see our project page at \url{gastwinformer.github.io}.
\end{abstract}
    
\vspace{-5pt}
\section{Introduction}
Methane (CH$_4$) represents a potent greenhouse gas with a global warming potential 84 times greater than carbon dioxide over a 20-year timeframe~\cite{pachauri2014climate}. Agriculture accounts for 40\% of human-caused methane emissions, with livestock responsible for roughly 32\%~\cite{shindell2021global}. As global food demand is expected to increase by 70\% by 2050, developing efficient monitoring systems for livestock methane emissions has become critical for climate mitigation~\cite{searchinger2018creating}.

The relationship between livestock diet composition and methane production creates opportunities for integrated monitoring systems. Different feed regimens significantly influence emission patterns—high-forage diets typically increase methane production due to fiber fermentation, while grain-rich diets can reduce emissions~\cite{embaby2025optical}. 
Advanced monitoring technologies enable precise quantification of these emission patterns, creating opportunities for data-driven livestock management through real-time assessment of feeding strategies and emission mitigation interventions~\cite{o2024advancements}.

Traditional methane quantification methods rely on respiration chambers or emission factor calculations, which suffer from high costs, labor-intensive protocols, and inability to capture real-time dynamics~\cite{tedeschi2022quantification}. Recent advances in optical gas imaging (OGI) offer non-invasive, continuous monitoring capabilities using thermal infrared cameras operating in the 7-8.5 $\mu$m spectral range~\cite{wang2024large,wang2020machine}. However, OGI presents computational challenges including low signal-to-noise ratios, complex thermal backgrounds, and irregular plume morphology requiring automated analysis~\cite{jahan2024deep}.

Vision transformers have revolutionized dense prediction tasks through global context modeling, but face computational challenges with high-resolution OGI data due to quadratic attention complexity~\cite{liang2022expediting}. Recent hybrid attention mechanisms show promise for balancing efficiency with representational capacity, but have not been adapted for gas plume segmentation~\cite{huang2019interlaced,li2021local,yang2021transformer}.

\begin{figure*}[t]
    \centering
    \captionsetup{belowskip=5pt}
    \includegraphics[width=1\linewidth]{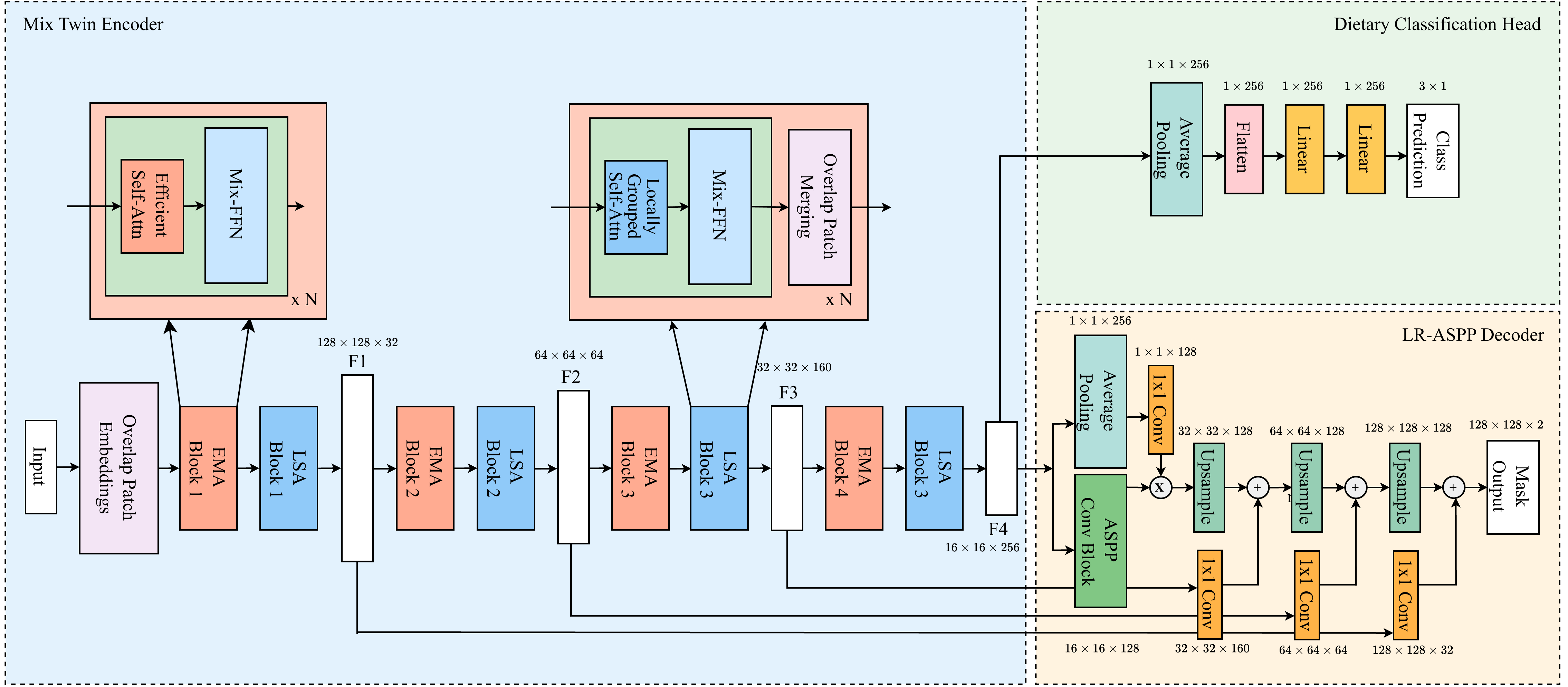}
    \caption{GasTwinFormer architecture. The Mix Twin encoder uses alternating EMA and LSA blocks across four hierarchical stages. The LR-ASPP decoder performs methane segmentation while a separate classification head predicts dietary treatment.}

    \label{fig:architecture}
    \vspace{-20pt}
\end{figure*}

We propose a novel architecture called GasTwinFormer for semantic segmentation of methane emissions and dietary treatment classification in beef cattle OGI camera images. In the encoding stage, we develop a Mix Twin encoder that combines efficient multi-head attention (EMA)~\cite{xie2021segformer} with locally-grouped self-attention (LSA)~\cite{chu2021twins} to capture both global context and local details for precise gas plume detection. This dual attention approach enables effective processing of thermal infrared imagery while maintaining computational efficiency. In the decoding stage, we use a hierarchical LR-ASPP decoder~\cite{howard2019searching} that processes features from multiple encoder stages to generate accurate segmentation predictions. Our framework performs both pixel-wise methane segmentation and dietary classification using shared features. The main contributions of this study are as follows:
\begin{enumerate}
    \item GasTwinFormer, a hybrid transformer-based architecture that enables concurrent methane plume segmentation and dietary treatment classification in livestock monitoring applications.
    \item A comprehensive beef cattle methane emission dataset captured through OGI technology, comprising 11,694 semi-automatically annotated frames spanning three feeding regimens.
    \item Extensive benchmarking and performance analysis against existing state-of-the-art segmentation approaches, demonstrating superior accuracy and computational efficiency across multiple metrics.
\end{enumerate}

\section{Related Work}

\noindent \textbf{Optical Gas Imaging and Methane Detection.} Wang \textit{et al.}~\cite{wang2020machine} pioneered computer vision for methane detection using infrared cameras, developing GasNet with 95\% detection accuracy on $\sim$1M labeled frames. VideoGasNet~\cite{wang2022videogasnet} extended this work to leak size classification using 3D CNNs. Recent advances include vision transformers for satellite methane detection~\cite{rouet2024automatic} and CNNs for airborne emission quantification~\cite{jahan2024deep}. Most recently, Sarker \textit{et al.}~\cite{sarker2024gasformer} introduced Gasformer, achieving 88.56\% mIoU on livestock datasets using Mix Vision Transformer encoders. However, existing approaches lack systematic integration of global and local attention for enhanced boundary delineation in challenging thermal imagery.

\noindent \textbf{Vision Transformers for Dense Prediction.} Dosovitskiy \textit{et al.}~\cite{dosovitskiy2020image} established Vision Transformers for image classification, while Ranftl \textit{et al.}~\cite{ranftl2021vision} introduced Dense Vision Transformers for dense prediction tasks. Hierarchical designs have proven effective: Swin Transformer~\cite{liu2021swin} uses shifted windowing for computational efficiency, PVT~\cite{wang2021pyramid} establishes hierarchical principles through progressive spatial reduction, and SegFormer~\cite{xie2021segformer} achieves state-of-the-art performance (51.8\% mIoU on ADE20K) through efficient MLP decoders and Mix Vision Transformers with spatial inductive bias.

\noindent \textbf{Hybrid Attention Mechanisms.} Chu \textit{et al.}~\cite{chu2021twins} proposed Twins architectures (PCPVT and SVT) that systematically combine different attention mechanisms. Twins introduces Locally-Grouped Self-Attention (LSA) partitioning spatial dimensions into non-overlapping windows for linear complexity, while maintaining local pattern recognition. 
Yang \textit{et al.}~\cite{yang2021transformer} demonstrated that treating global and local attention as complementary achieves superior dense prediction performance. However, existing hybrid approaches focus on natural images and have not been adapted for gas plume segmentation challenges.

\noindent \textbf{Research Gaps.} Despite advances in methane detection and vision transformers, three critical limitations hinder practical OGI-based livestock monitoring. Current limitations in OGI-based livestock monitoring include: \textbf{(1)} reliance on single-scale attention mechanisms that inadequately balance global context and local precision for gas plume characteristics, \textbf{(2)} treatment of methane detection as an isolated task without leveraging established diet-emission correlations, and \textbf{(3)} absence of comprehensive livestock-specific datasets capturing real-world farming complexities beyond controlled laboratory conditions. Our GasTwinFormer addresses these limitations through hybrid attention design, multi-task learning integration, and comprehensive dataset development.

\vspace{-5pt}
\section{Method}
GasTwinFormer consists of three primary components: (1) a hierarchical Mix Twin encoder that combines efficient multi-head self-attention (EMA) from SegFormer's Mix Transformer~\cite{xie2021segformer} with locally-grouped self-attention (LSA) from Twins~\cite{chu2021twins}, (2) a hierarchical lightweight reduced Atrous Spatial Pyramid Pooling decoder (LR-ASPP)~\cite{howard2019searching} for multi-scale feature aggregation and pixel-wise methane segmentation, and (3) a dietary classification head for scene-level prediction. ~\Cref{fig:architecture} illustrates the complete architecture pipeline.

\subsection{Mix Twin Encoder}

\noindent \textbf{Hybrid Attention Architecture.} The backbone encoder follows a hierarchical design with four stages that progressively reduce spatial resolution from $H/4$ to $H/32$ while expanding channel capacity from 32 to 256. Within each stage, we use an EMA$\rightarrow$LSA composition: an EMA block establishes global relationships via spatially reduced attention, followed by an LSA block that refines local structure using $5\times 5$ windows. Each stage therefore contains exactly one EMA--LSA pair (EL). 
SegFormer uses only efficient (EMA) attention in every block. Twins--PCPVT uses only global sub-sampled attention (GSA). Twins--SVT places LSA first and GSA second in a repeating LSA$\rightarrow$GSA sequence. 
Accordingly, composing EL within each stage captures long-range plume context via EMA and sharp boundary details via LSA in a single pass, yielding higher accuracy than EMA-only (SegFormer), GSA-only (Twins--PCPVT), or LSA$\rightarrow$GSA alternation (Twins--SVT); quantitative gains are reported in ~\cref{tab:comparison_results}.

\noindent \textbf{Hierarchical Multi-Scale Feature Extraction.} Given an input image $I \in \mathbb{R}^{H \times W \times 3}$, the encoder generates multi-scale feature representations $\{F_1, F_2, F_3, F_4\}$ with progressive spatial downsampling and corresponding channel dimensions $\{32, 64, 160, 256\}$, respectively. 
The EMA-LSA pairs in each stage incorporate overlapped patch embedding and layer normalization, resulting in 8 total blocks across the four-stage encoder.

\noindent \textbf{Overlapped Patch Embedding.} We use overlapped patch embedding to preserve spatial continuity for precise boundary localization. The first stage uses a $7 \times 7$ convolution with stride 4 and padding 3, while subsequent stages use $3 \times 3$ convolutions with stride 2 and padding 1 for efficient downsampling.

\noindent \textbf{Efficient Multi-Head Attention.} Standard multi-head self-attention mechanisms exhibit quadratic computational complexity $O(N^2)$ with respect to spatial resolution N = H × W, creating computational bottlenecks for high-resolution dense prediction tasks. We address this limitation by adopting the Efficient Multi-Head Attention (EMA) from SegFormer~\cite{xie2021segformer}, which builds upon the spatial reduction process introduced in Pyramid Vision Transformer~\cite{xie2021segformer}. This approach reduces complexity to $O(N^2/R)$ through spatial reduction of key and value representations while maintaining full-resolution queries.
For each stage $i$ with reduction ratio $R_i$, both key and value matrices are spatially downsampled to dimensions $\mathbb{R}^{(N/R_i) \times C}$ using convolutions with kernel size and stride equal to $R_i$. The attention computation becomes:

{\footnotesize
\vspace{-5pt}
\begin{equation}
\begin{aligned}
\text{Attention}(\mathbf{Q}, \mathbf{K}', \mathbf{V}') = \text{Softmax}\left(\frac{\mathbf{Q}(\mathbf{K}')^T}{\sqrt{d_{\text{head}}}}\right)\mathbf{V}'
\end{aligned}
\end{equation}
\vspace{-5pt}
}

where $\mathbf{K}'$ and $\mathbf{V}'$ are the spatially reduced key and value representations with dimensions $\mathbb{R}^{(N/R_i) \times C}$.

We use stage-adaptive reduction ratios $R = \{8, 4, 2, 1\}$ that align with the hierarchical nature of feature learning. Early stages utilize aggressive reduction ($R = 8$) to handle high-resolution features efficiently, while later stages progressively decrease reduction ratios as spatial dimensions naturally diminish through downsampling. This strategy ensures computational tractability in high-resolution stages while maintaining fine-grained attention capabilities in semantically rich later stages.

\noindent \textbf{Locally-Grouped Self-Attention.} While EMA achieves computational efficiency through spatial reduction, it may compromise fine-grained spatial detail preservation that is critical for accurate boundary delineation in methane plume segmentation. To address this limitation, we integrate LSA from Twins-SVT~\cite{chu2021twins} as the second component in our hybrid attention pattern. LSA complements the global context modeling of efficient attention by capturing fine-grained local structures through spatially partitioned attention computation.

The LSA addresses the quadratic complexity challenge through spatial partitioning rather than spatial reduction. Given an input feature map $\mathbf{X} \in \mathbb{R}^{B \times N \times C}$ where $N = H \times W$, LSA partitions the spatial dimensions into non-overlapping windows of size $w_1 \times w_2$. Self-attention is then computed independently within each local window:

{\footnotesize
\vspace{-5pt}
\begin{equation}
\begin{aligned}
\text{LSA}(\mathbf{X}) = \text{Concat}_{i,j}\left(\text{Attention}(\mathbf{X}_{i,j})\right)
\end{aligned}
\end{equation}
\vspace{-10pt}
}

where $\mathbf{X}_{i,j} \in \mathbb{R}^{B \times w_1w_2 \times C}$ represents the feature tokens within window $(i,j)$, and the concatenation operates over all $\lceil H/w_1 \rceil \times \lceil W/w_2 \rceil$ windows. The attention computation within each window follows the standard formulation:

{\footnotesize
\vspace{-5pt}
\begin{equation}
\begin{aligned}
\text{Attention}(\mathbf{X}_{i,j}) = \text{Softmax}\left(\frac{\mathbf{Q}_{i,j}\mathbf{K}_{i,j}^T}{\sqrt{d_{\text{head}}}}\right)\mathbf{V}_{i,j}
\end{aligned}
\end{equation}
\vspace{-5pt}
}

This design achieves computational complexity of $\mathcal{O}(w_1w_2HWd)$, which scales linearly with spatial resolution since the window size $w_1w_2$ remains fixed. For our implementation with $w_1 = w_2 = 5$, the complexity becomes $\mathcal{O}(25HWd)$, providing substantial efficiency gains while maintaining sufficient receptive field coverage for local pattern recognition.

\noindent \textbf{Mix Feed-Forward Network.} Both Transformer Block and LSA Block utilize the Mix Feed-Forward Network (Mix-FFN) module from SegFormer~\cite{xie2021segformer}, which eliminates the need for explicit positional encodings while providing spatial inductive bias. Unlike Vision Transformers that use fixed-resolution positional encodings, we argue that positional encoding is not necessary for dense prediction tasks. Instead, Mix-FFN considers the effect of zero padding to leak location information by directly incorporating a $3 \times 3$ convolution in the feed-forward network.
The Mix-FFN operation is formulated as:

{\footnotesize
\vspace{-5pt}
\begin{equation}
\begin{aligned}
\text{Mix-FFN}(\mathbf{x}) = \text{MLP}(\text{GELU}(\text{Conv}_{3 \times 3}(\text{MLP}(\mathbf{x})))) + \mathbf{x}
\end{aligned}
\end{equation}
\vspace{-15pt}
}

where $\mathbf{x}$ is the feature from the self-attention module. 
Mix-FFN mixes a $3 \times 3$ convolution and MLPs into each feed-forward network. The $3 \times 3$ convolution is sufficient to provide positional information for transformers through the spatial connectivity and zero-padding effects. We use depth-wise convolutions to reduce the number of parameters and improve computational efficiency.

\subsection{Hierarchical LR-ASPP Decoder}
\label{sec:lraspp}
For dense prediction tasks, we use a lightweight decoder that efficiently aggregates multi-scale features from our hierarchical encoder. Building upon LR-ASPP from MobileNetV3~\cite{howard2019searching}, we propose an adaptive variant that accommodates variable input resolutions while maintaining computational efficiency.
Our Hierarchical LR-ASPP decoder processes multi-scale features $\{F_1, F_2, F_3, F_4\}$ through two parallel pathways: $F_4$ features are processed through the main ASPP path, while $F_1, F_2, F_3$ features are processed through dedicated 1$\times$1 convolution branches. The operations are:

{\footnotesize
\vspace{-5pt}
\begin{equation}
\begin{aligned}
\makebox[3em][l]{$F_{\text{pool}}$} &= \text{Sigmoid}(\text{Conv}_{1 \times 1}(\text{AdaptiveAvgPool}(F_4))) \\
\makebox[3em][l]{$F_{\text{aspp}}$} &= \text{Conv}_{1 \times 1}(F_4) \odot \text{Upsample}(F_{\text{pool}}) \\
\makebox[3em][l]{$F_{\text{branch}_i}$} &= \text{Conv}_{1 \times 1}(F_i), \quad i \in \{1, 2, 3\} \\
\makebox[3em][l]{$F_{\text{out}}$} &= \text{ProgressiveFusion}(F_{\text{aspp}}, \{F_{\text{branch}_3}, F_{\text{branch}_2}, F_{\text{branch}_1}\})
\end{aligned}
\end{equation}
\vspace{-5pt}
}

where $\odot$ denotes element-wise multiplication, and progressive fusion sequentially upsamples, concatenates, and fuses features from deeper to shallower levels. 
This design preserves both semantic information from deep features and spatial details from shallow features essential for accurate methane plume boundary delineation.


\subsection{Dietary Classification Head}

To enable simultaneous scene-level classification alongside dense plume segmentation, we incorporate a lightweight classification head that processes the highest-level semantic features from the encoder. The classification head employs a simple yet effective architecture consisting of adaptive average pooling, followed by a two-layer fully connected network with ReLU activation and dropout regularization.
The classification head operates on the final stage features $F_4$ to predict dietary treatment categories: High Forage (HF), Mixed Diet (MD), and High Grain (HG).

\subsection{Gaussian Plume Weighted Dice Loss}
\label{sec:gaussian_loss}

We incorporate the Gaussian Plume Weighted Dice Loss~\cite{zhou13high} to leverage physical constraints from gas dispersion behavior in our segmentation framework. This loss function addresses the inherent characteristics of gas plume dynamics by applying spatially-varying weights based on the Gaussian plume model. The loss formulation applies pixel-wise weights according to the Gaussian distribution:

{\footnotesize
\vspace{-5pt}
\begin{equation}
\begin{aligned}
w(p) = \exp\left(-\frac{(p_x-\mu_x)^2}{2\sigma_x^2} - \frac{(p_y-\mu_y)^2}{2\sigma_y^2}\right)
\end{aligned}
\end{equation}
\vspace{-5pt}
}

where $p = (p_x, p_y)$ denotes pixel coordinates, $(\mu_x, \mu_y)$ represents the plume center computed via center-of-mass on predicted masks, and $(\sigma_x, \sigma_y)$ denote the horizontal and vertical diffusion scales estimated through weighted standard deviation with adaptive bounds $[\frac{W}{20}, \frac{W}{2}]$ and $[\frac{H}{20}, \frac{H}{2}]$ respectively, where $W$ and $H$ are the image dimensions. The weighted Dice loss is then computed as:

{\footnotesize
\vspace{-5pt}
\begin{equation}
\begin{aligned}
L_{\text{weighted}} = 1 - \frac{2\sum_{p} w(p) \cdot y_p \cdot \hat{y}_p + \epsilon}{\sum_{p} w(p) \cdot y_p + \sum_{p} w(p) \cdot \hat{y}_p + \epsilon}
\end{aligned}
\end{equation}
\vspace{-5pt}
}

where $y_p$ and $\hat{y}_p$ represent the ground truth and predicted segmentation values at pixel $p$, and $\epsilon$ is a small constant for numerical stability.

\vspace{-5pt}
\section{Beef Cattle Methane Emission Dataset}

We present a comprehensive dataset for methane emission detection from beef cattle, captured using this OGI camera. This dataset addresses the critical need for computer vision benchmarks in livestock emission monitoring, particularly for developing and evaluating segmentation algorithms under challenging real-world conditions. We use the FLIR Gx320 OGI camera for methane emission detection. The camera operates in the 3.2--3.4~$\mu$m spectral range, optimized for hydrocarbon detection through mid-wave infrared sensing. Key specifications include 320$\times$240 pixel resolution and $<$10~mK thermal sensitivity. The camera detects methane concentrations as low as 9.6~ppm$\cdot$m under optimal conditions with 10$^\circ$C thermal contrast~\cite{flir_gseries_datasheet}.

\noindent \textbf{In Vivo Trial Design.} The primary aim of this in vivo trial was to assess the efficacy of combining optical gas imaging with deep learning to detect and segment methane emissions from ruminant animals across different dietary treatments. The study utilized twelve postpartum beef cows (1200 lb ± 23) over a 30-day period, with 4 animals assigned to one of three dietary treatment groups. Each group was housed and fed together in separate feed stalls, receiving 30 lb of diet mix per cow daily. All cows received feed twice daily at 7 AM and 7 PM, where hay was offered first, followed by the grain mix. All animals had free access to clean water and were housed at Southern Illinois University's beef center barns, managed in accordance with Institutional Animal Care and Use Committee guidelines (protocol number 22-016)~\cite{olaw2002iacuc}.
We conducted controlled experiments across three dietary treatment groups to investigate the relationship between feed composition and methane emissions: High Forage Group (HF) fed 100\% hay consisting of grass and legume mix;
Mixed Diet Group (MD) fed 50\% hay mix and 50\% grain mix (67.5\% corn, 25\% DDGS, and 7.5\% mineral mix); and High Grain Group (HG) fed 20\% hay mix and 80\% grain mix, with grain levels increased gradually to prevent acidosis and facilitate adaptation to the high-grain diet. Every cow was kept in an animal chute for 20 minutes for gas recording, and at the end of the experiment, all cows were moved to the holding barn two hours after morning feeding. The gas imaging was performed using the TELEDYNE FLIR Gx320, with the infrared camera mounted in a lateral position approximately 4 feet from the cow's head. Following recording, cows were returned to their assigned barn. The same recording procedure was repeated the next day to collect additional data required for model training.

\noindent \textbf{Image Acquisition.} Videos were captured in black-hot thermal mode (dark gas plumes against light backgrounds) in FLIR's CSQ format with 14-bit radiometric data.
We converted CSQ files to MP4 using FLIR Thermal Studio, then extracted frames as 8-bit grayscale PNG images (0-255 intensity values with three identical channels).

\begin{table}[tpb]
\centering
\scriptsize
\begin{tabular}{@{}lcccccc@{}}
\toprule
\textbf{Diet Type} & \textbf{Images} & \textbf{Percentage} & \textbf{Videos} & \textbf{\makecell{Train\\(70\%)}} & \textbf{\makecell{Val\\(15\%)}} & \textbf{\makecell{Test\\(15\%)}} \\
\midrule
High Forage & 2,730 & 23.4\% & 10 & 1,906 & 404 & 420 \\
Mixed Diet & 4,658 & 39.8\% & 5 & 3,258 & 696 & 704 \\
High Grain & 4,306 & 36.8\% & 4 & 3,013 & 644 & 649 \\
\midrule
\textbf{Total} & \textbf{11,694} & \textbf{100.0\%} & \textbf{19} & \textbf{8,177} & \textbf{1,744} & \textbf{1,773} \\
\bottomrule
\end{tabular}
\vspace{-8pt}
\caption{Beef cattle methane emission dataset statistics. Percentages show distribution within 11,694 annotated frames.}
\label{tab:dataset_comprehensive}
\vspace{-15pt}
\end{table}

\noindent \textbf{Dataset Statistics and Composition.} Our dataset comprises 208,149 frames extracted at 30 fps from 19 FLIR thermal recordings across dietary treatment groups. Each frame has 640$\times$480 pixel resolution and is stored as 8-bit grayscale PNG files with values ranging 0-255.
We identified and annotated 11,694 frames (5.6\% of 208,149 frames) containing visible methane plumes, reflecting the intermittent nature of bovine eructation events. As shown in ~\cref{tab:dataset_comprehensive}, the annotated frames are distributed across dietary treatments as: 4,658 mixed diet (39.8\%), 4,306 high grain (36.8\%), and 2,730 high forage (23.4\%) frames. This distribution reflects both biological emission differences and collection constraints across treatments.
For model development, we employed temporal splitting to preserve emission sequence integrity: 70\% of consecutive frames for training, 15\% for validation, and 15\% for testing within each video. ~\Cref{tab:dataset_comprehensive} details the resulting splits: 8,177 training frames, 1,744 validation frames, and 1,773 test frames. This ensures evaluation on future time points relative to training data, providing realistic generalization assessment. All 19 videos contribute to each split while maintaining dietary treatment proportions. We excluded the remaining $\sim$196k non-plume frames to avoid severe class imbalance without meaningful segmentation training signal.

\noindent \textbf{Annotation Methodology.} We developed a multi-stage annotation pipeline combining classical image processing, deep learning, and manual refinement to generate reliable ground truth masks for ephemeral methane plumes with low contrast and irregular morphology. 
Our pipeline consists of three complementary approaches: (1) \textit{Classical processing} employs temporal background subtraction using exponential moving average over 5 frames, followed by motion masking (thresholds 20-60), adaptive mean thresholding (block sizes 300-5001, constants 5-15), watershed segmentation with Sobel edge detection, and morphological refinement with size filtering ($>$2000 pixels) and eccentricity filtering ($>$0.95) to remove linear artifacts. (2) \textit{Deep learning processing} uses a Gasformer~\cite{sarker2024gasformer} model trained on initial classical masks to identify subtle patterns beyond traditional methods. (3) \textit{Enhanced processing} applies CLAHE, intensity rescaling, and non-local means denoising ($h = 15$) for improved candidate generation.
For each frame, we generate three mask candidates from these approaches and perform manual inspection to select the most accurate representation using contrast-enhanced overlays. 
~\Cref{fig:qualitative_comparison} (second column) shows the resulting ground truth masks for all three dietary treatments.

\vspace{-5pt}
\section{Results}

\begin{table*}[tpb]
\centering
\scriptsize
\begin{tabular}{@{}l|l|cccc|cc|c|c@{}}
\toprule
\textbf{Method} & \textbf{Backbone} & \textbf{\makecell{mIoU\\(\%)$\uparrow$}} & \textbf{\makecell{mF1\\(\%)$\uparrow$}} & \textbf{\makecell{Diet Acc\\(\%)$\uparrow$}} & \textbf{\makecell{Diet F1\\(\%)$\uparrow$}} & \textbf{\makecell{Params\\(M)$\downarrow$}} & \textbf{\makecell{FLOPs\\(G)$\downarrow$}} & \textbf{\makecell{FPS\\$\uparrow$}} & \textbf{Year} \\
\midrule
\multicolumn{10}{l}{\textit{Transformer-based Methods}} \\
\midrule
SegFormer~\cite{xie2021segformer} & MiT-B0 & 72.11 & 81.57 & 100.0 & 100.0 & 3.782 & 7.885 & 119.66 & 2021 \\
Twins~\cite{chu2021twins} & PCPVT-S & 74.05 & 83.25 & 100.0 & 100.0 & 27.906 & 44.34 & 61.60 & 2021 \\
Twins~\cite{chu2021twins} & SVT-S & 72.06 & 81.62 & 100.0 & 100.0 & 27.846 & 38.471 & 51.64 & 2021 \\
GasFormer~\cite{sarker2024gasformer} & MiT-B0 & 72.25 & 81.69 & 100.0 & 100.0 & 3.716 & 9.913 & 102.29 & 2024 \\
iFormer~\cite{zheng2025iformer} & iFormer-T & 65.99 & 75.87 & 99.77 & 99.80 & 6.804 & 24.267 & 113.83 & 2025 \\

\midrule
\multicolumn{10}{l}{\textit{CNN-based Methods}} \\
\midrule
DeepLabV3~\cite{chen2017rethinking} & ResNet-50 & 70.36 & 80.03 & 100.0 & 100.0 & 68.625 & 270.0 & 91.79 & 2017 \\
BiSeNetV1~\cite{yu2018bisenet} & ResNet-18 & 52.87 & 59.29 & 95.32 & 95.76 & 13.455 & 14.821 & \textbf{243.07} & 2018 \\
Fast-FCNN~\cite{poudel2019fast} & FastSCNN & 54.01 & 61.09 & 97.01 & 96.95 & \textbf{1.488} & \textbf{0.927} & 225.79 & 2019 \\
ICNet~\cite{zhao2018icnet} & ResNet-50 & 63.40 & 73.04 & 100.0 & 100.0 & 47.859 & 15.426 & 138.55 & 2018 \\
UperNet~\cite{xiao2018unified} & ResNet-50 & 70.67 & 80.32 & 100.0 & 100.0 & 66.927 & 237.0 & 85.06 & 2018 \\
BiSeNetV2~\cite{yu2021bisenet} & BiSeNetV2 & 66.53 & 76.32 & 98.08 & 98.28 & 14.821 & 12.286 & 172.48 & 2021 \\
DDRNet~\cite{pan2022deep} & DDRNet & 68.91 & 78.65 & 99.94 & 99.94 & 5.766 & 4.56 & 156.38 & 2022 \\
RepViT~\cite{wang2024repvit} & RepViT-M0.9 & 68.03 & 77.93 & 100.0 & 100.0 & 8.954 & 25.404 & 84.30 & 2024 \\
\midrule
\textbf{GasTwinFormer} & \textbf{MixTwinEncoder} & \textbf{74.47} & \textbf{83.63} & \textbf{100.0} & \textbf{100.0} & 3.348 & 3.428 & 114.9 & 2025 \\
\bottomrule
\end{tabular}
\vspace{-5pt}
\caption{Comparison with state-of-the-art methods on our beef cattle methane emission dataset. $\uparrow$ indicates higher is better, $\downarrow$ indicates lower is better. Bold indicates better}
\label{tab:comparison_results}
\vspace{-15pt}

\end{table*}

\subsection{Implementation Details}

We implement all experiments using PyTorch and MMSegmentation framework on a server with Intel Xeon Gold 6338 (2.00GHz), NVIDIA A100 80GB GPU, and 512GB RAM. We evaluate GasTwinFormer against comprehensive baselines spanning transformer-based architectures (SegFormer~\cite{xie2021segformer}, Twins PCPVT-S~\cite{chu2021twins}, Twins SVT-S~\cite{chu2021twins}, Gasformer~\cite{sarker2024gasformer}, iFormer~\cite{zheng2025iformer}) and CNN-based methods (DeepLabV3~\cite{chen2017rethinking}, BiSeNetV1~\cite{yu2018bisenet}, Fast-FCNN~\cite{poudel2019fast}, ICNet~\cite{zhao2018icnet}, UperNet~\cite{xiao2018unified}, BiSeNetV2~\cite{yu2021bisenet}, DDRNet~\cite{pan2022deep}, RepViT~\cite{wang2024repvit}), with all models utilizing ImageNet pre-trained weights where available. For the main results, GasTwinFormer uses the EL-EL-EL-EL hybrid attention pattern with 5×5 LSA window size and Gaussian Plume Weighted Dice Loss, as determined optimal through ablation studies in ~\cref{sec:ablation}. Training proceeds for 80,000 iterations using AdamW optimizer (learning rate $6 \times 10^{-5}$, $\beta_1=0.9$, $\beta_2=0.999$, weight decay 0.01) with linear warmup from $10^{-6}$ over 1,500 iterations followed by polynomial decay (power=1.0). For GasTwinFormer, we initialize compatible components (patch embeddings, efficient attention, feed-forward networks) from SegFormer pre-trained weights while LSA layers are randomly initialized due to architectural novelty, using $10\times$ learning rate scaling for the decoder head and zero weight decay for normalization layers. Input images are resized to $512 \times 512$ pixels with data augmentation including random horizontal flipping (50\% probability) and photometric distortion, using batch size 8 for training, and batch size 1 for inference. The multi-task pipeline handles simultaneous segmentation and classification annotations with validation every 8,000 iterations, retaining the top 3 checkpoints based on mean IoU performance. We report segmentation performance using mean Intersection over Union (mIoU) and mean F1-score (mF1), classification performance using accuracy and F1-score, and computational efficiency via parameters, FLOPs, and inference speed (FPS), with all metrics computed on the test set using the best validation checkpoint.

\vspace{-5pt}
\begin{figure*}
    \centering

    \captionsetup[subfloat]{labelformat=empty}

    \rotatebox[origin=c]{90}{\parbox{0cm}{\centering MD}}\hspace{5pt}%
    \subfloat{\includegraphics[width=0.28\columnwidth]{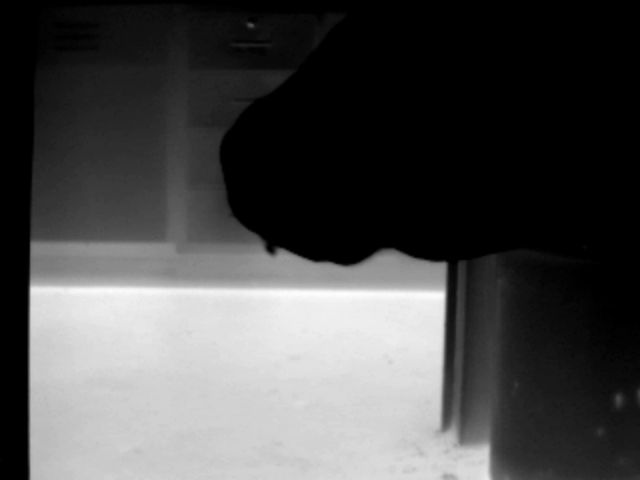}} \hspace{0pt}
    \subfloat{\includegraphics[width=0.28\columnwidth]{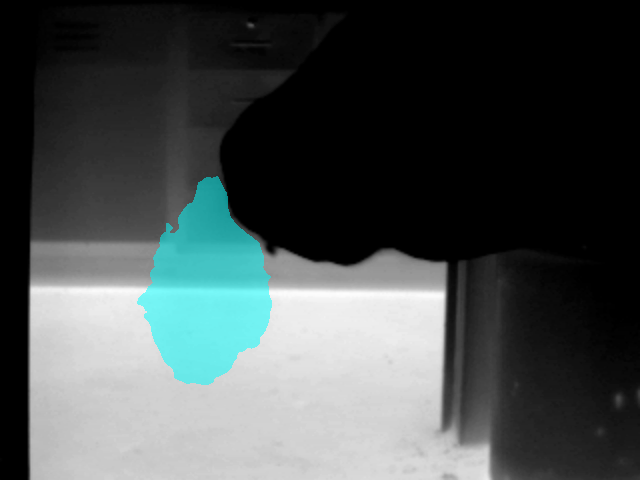}} \hspace{0pt}
    \subfloat{\includegraphics[width=0.28\columnwidth]{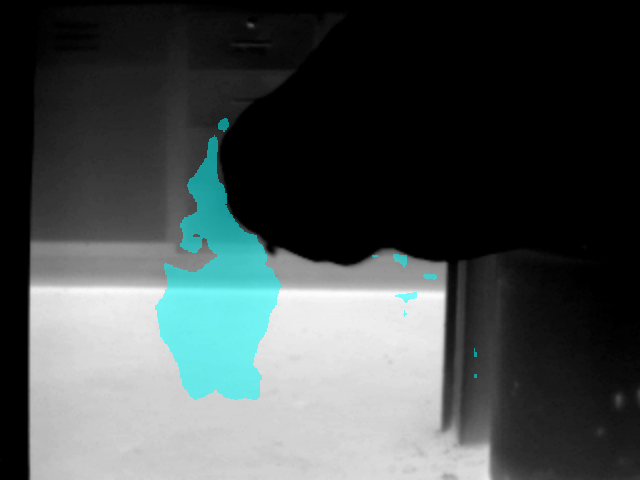}} \hspace{0pt}
    \subfloat{\includegraphics[width=0.28\columnwidth]{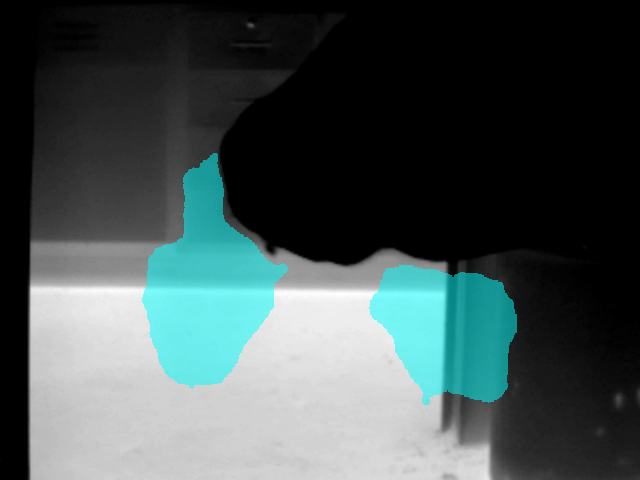}} \hspace{0pt}
    \subfloat{\includegraphics[width=0.28\columnwidth]{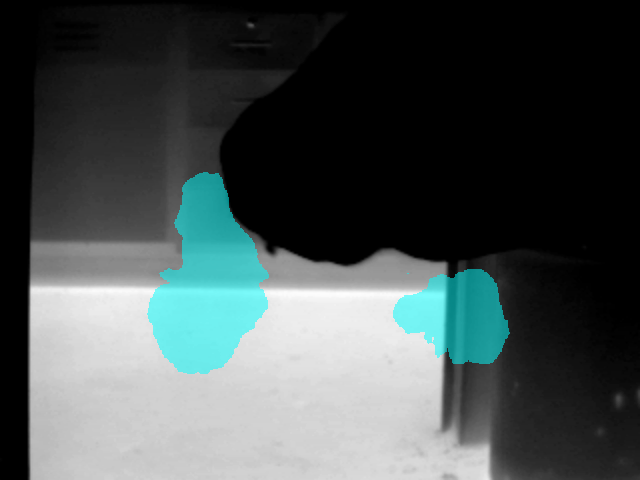}} \hspace{0pt}
    \subfloat{\includegraphics[width=0.28\columnwidth]{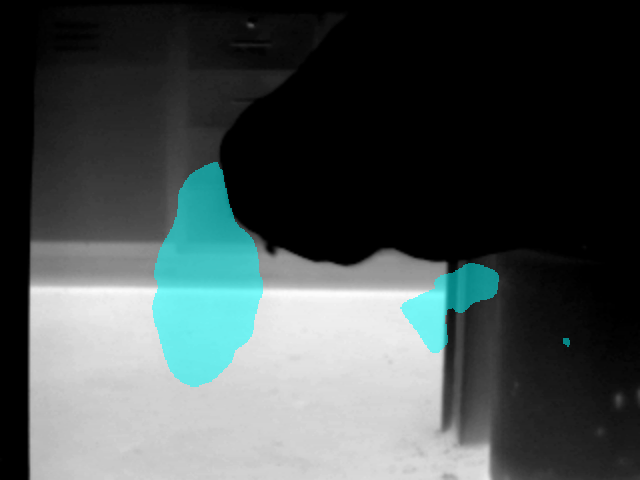}} \hspace{0pt}
    \subfloat{\includegraphics[width=0.28\columnwidth]{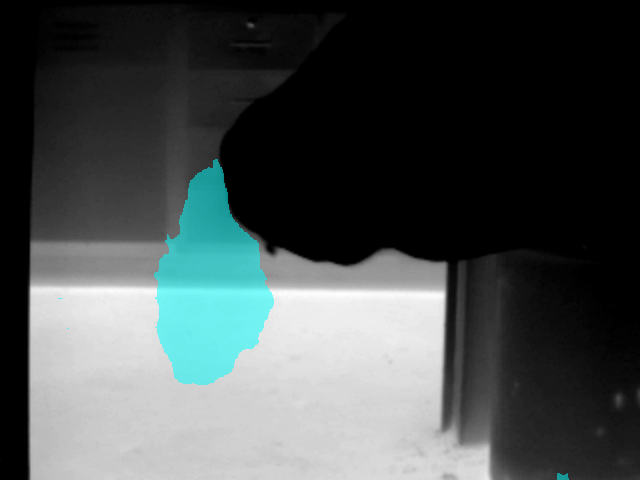}}  \\

    \rotatebox[origin=c]{90}{\parbox{0cm}{\centering HG}}\hspace{5pt}%
    \subfloat{\includegraphics[width=0.28\columnwidth]{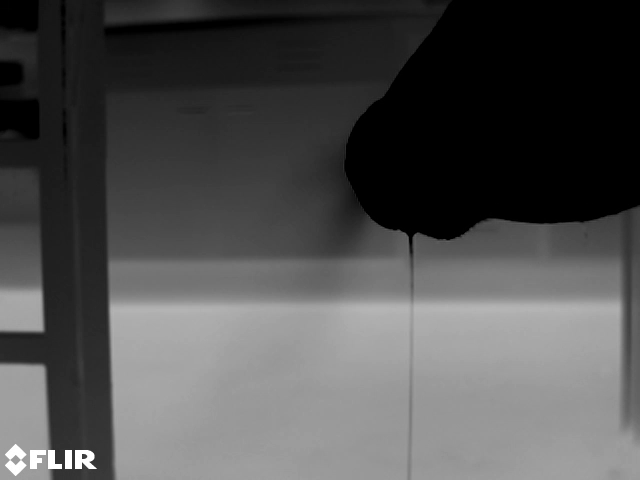}} \hspace{0pt}
    \subfloat{\includegraphics[width=0.28\columnwidth]{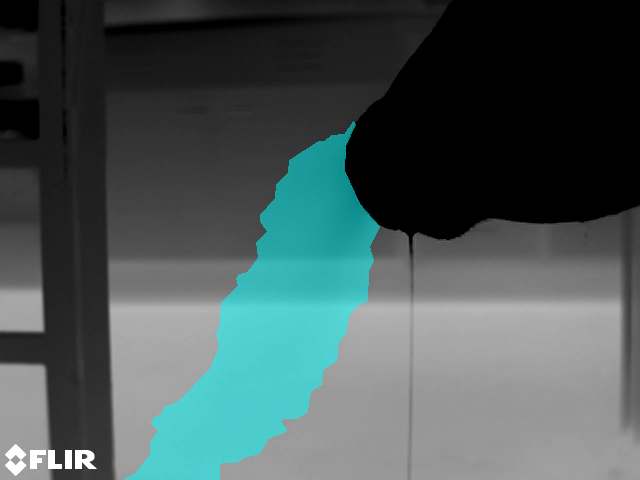}} \hspace{0pt}
    \subfloat{\includegraphics[width=0.28\columnwidth]{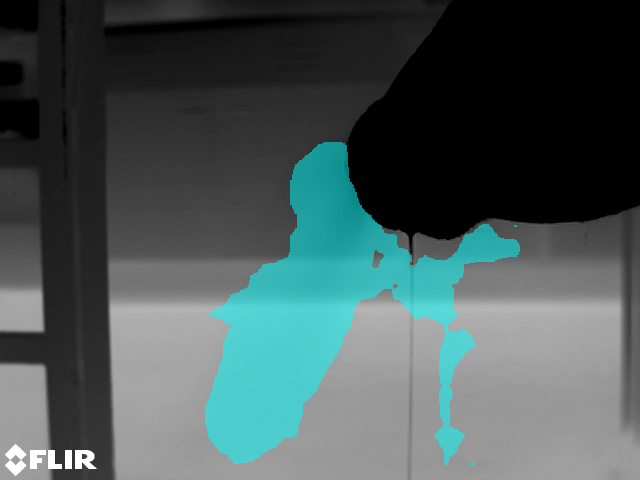}} \hspace{0pt}
    \subfloat{\includegraphics[width=0.28\columnwidth]{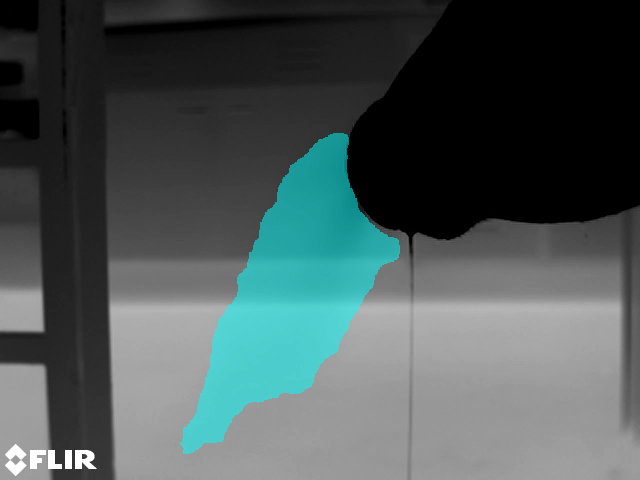}} \hspace{0pt}
    \subfloat{\includegraphics[width=0.28\columnwidth]{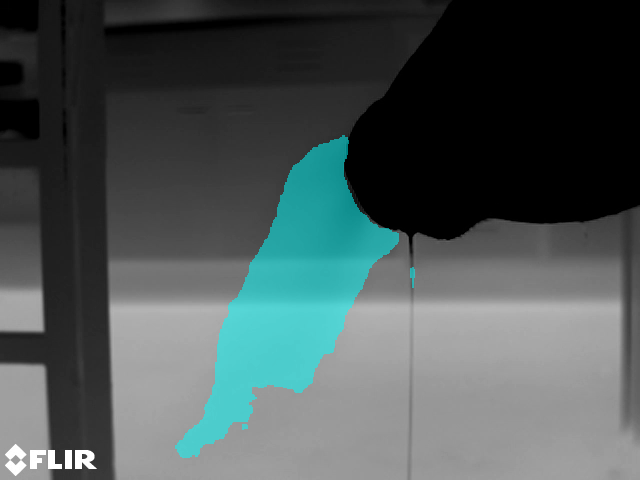}} \hspace{0pt}
    \subfloat{\includegraphics[width=0.28\columnwidth]{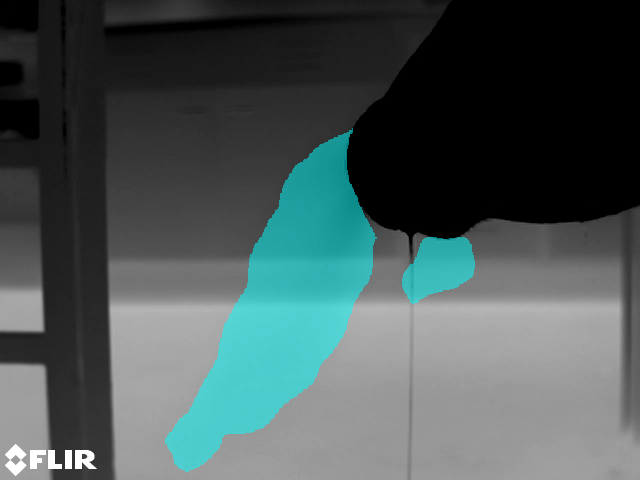}} \hspace{0pt}
    \subfloat{\includegraphics[width=0.28\columnwidth]{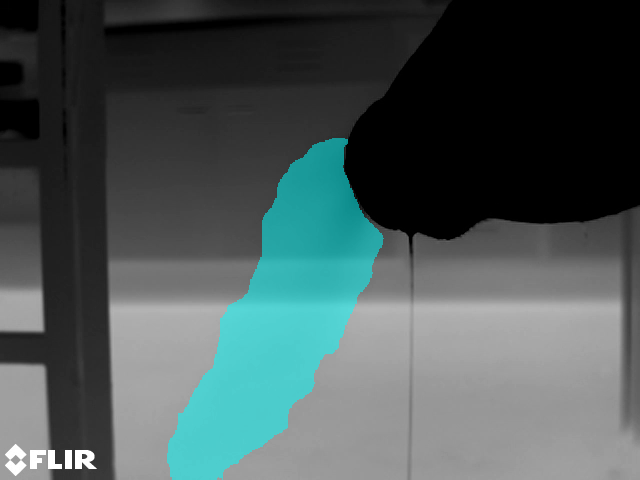}}  \\

    \rotatebox[origin=c]{90}{\parbox{0cm}{\centering HF}}\hspace{5pt}%
    \subfloat[Image]{\includegraphics[width=0.28\columnwidth]{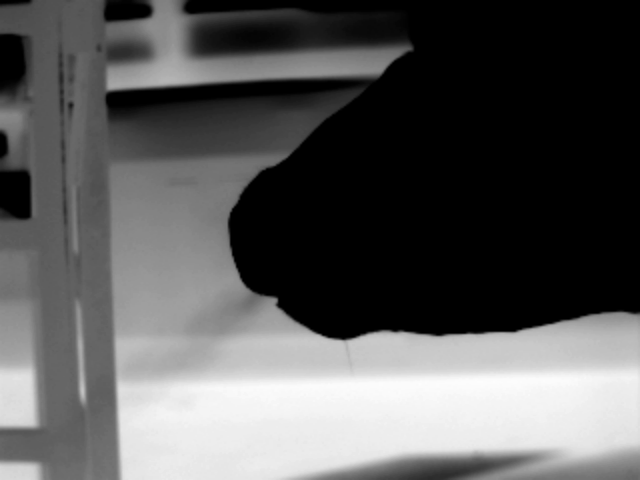}} \hspace{0pt}
    \subfloat[Ground Truth]{\includegraphics[width=0.28\columnwidth]{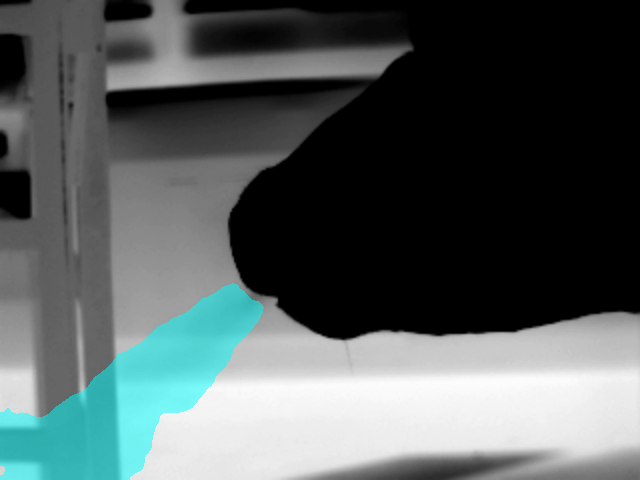}} \hspace{0pt}
    \subfloat[BiSeNetV2]{\includegraphics[width=0.28\columnwidth]{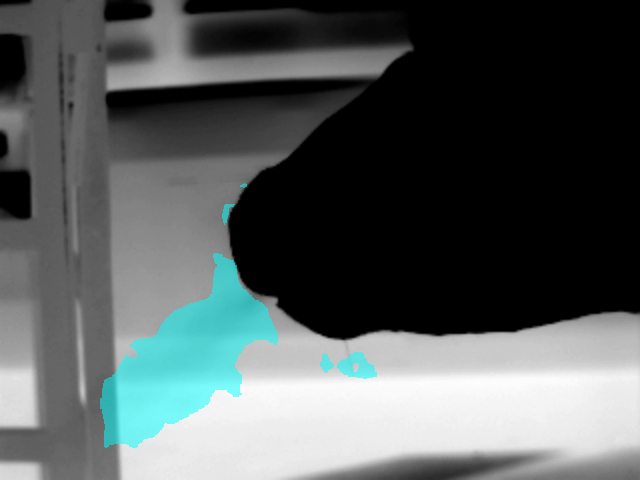}} \hspace{0pt}
    \subfloat[GasFormer]{\includegraphics[width=0.28\columnwidth]{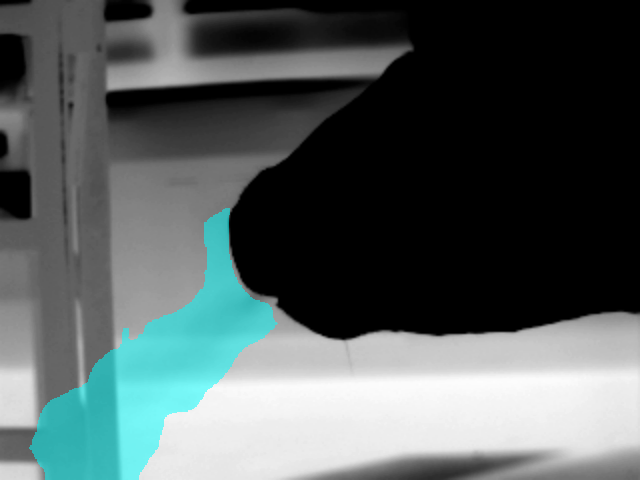}} \hspace{0pt}
    \subfloat[Segformer]{\includegraphics[width=0.28\columnwidth]{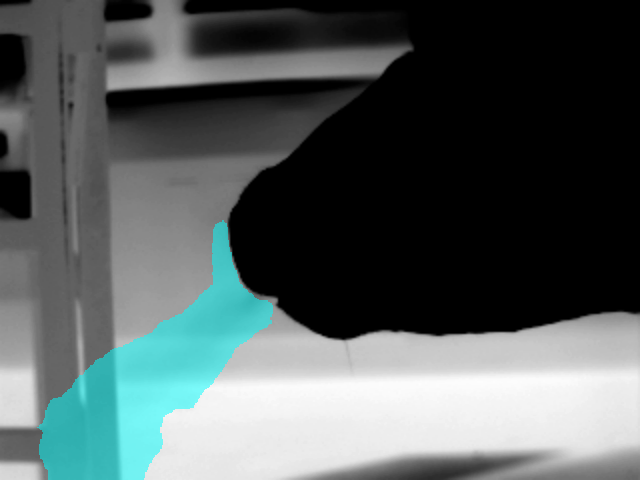}} \hspace{0pt}
    \subfloat[Twins PCPVT-S ]{\includegraphics[width=0.28\columnwidth]{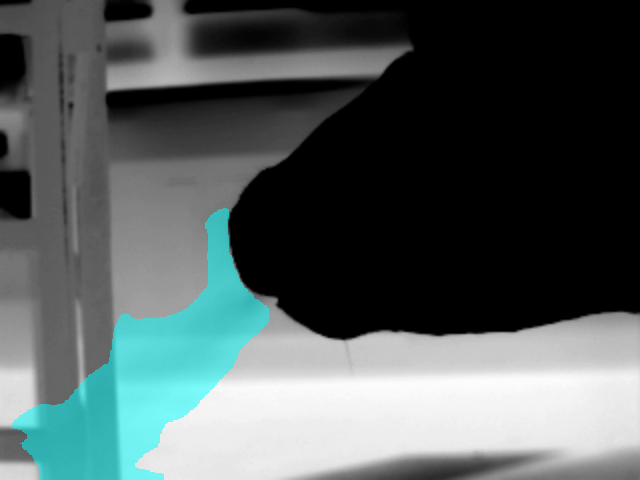}} \hspace{0pt}
    \subfloat[GasTwinFormer]{\includegraphics[width=0.28\columnwidth]{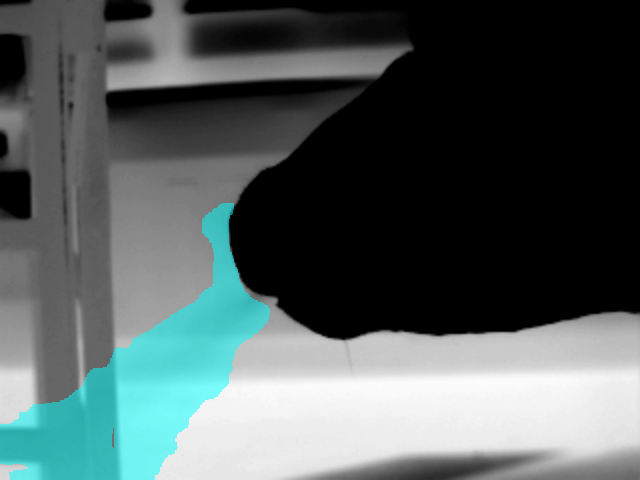}}  \\

    \vspace{-5pt}
    \caption{Qualitative comparison of methane plume segmentation results across different models and dietary treatments (MD: mixed diet, HG: high grain, HF: high forage), with ground truth masks shown for reference.}
    \label{fig:qualitative_comparison}
    \vspace{-15pt}
\end{figure*}

\subsection{Comparison with state-of-the-arts}

We evaluate GasTwinFormer against transformer-based and CNN-based methods on our beef cattle methane emission dataset. ~\Cref{tab:comparison_results} summarizes performance metrics for segmentation and dietary classification tasks.

\noindent \textbf{Segmentation Performance Analysis.} GasTwinFormer achieves $74.47\%$ mIoU and $83.63\%$ mF1 using only $3.348$M parameters and $3.428$G FLOPs, outperforming all other approaches in terms of accuracy while maintaining exceptional efficiency. For instance, compared to Gasformer, GasTwinFormer delivers $2.22\%$ better mIoU while requiring $9.9\%$ fewer parameters and $65.4\%$ fewer FLOPs. Compared to SegFormer, GasTwinFormer achieves $2.36\%$ better mIoU and $2.06\%$ better mF1 while requiring $11.5\%$ fewer parameters and $56.5\%$ fewer FLOPs. Moreover, GasTwinFormer outperforms all transformer-based approaches, including Twins PCPVT-s, achieving $0.42\%$ better mIoU while being significantly more efficient with $8.3\times$ fewer parameters and $12.9\times$ fewer FLOPs.
Compared to heavyweight CNN methods, our results demonstrate substantial superiority. Our method represents a $3.8\%$ improvement over UperNet and a $4.11\%$ improvement over DeepLabV3, while requiring $20\times$ fewer parameters and running $69$--$78\times$ more efficiently in terms of FLOPs. Among efficient approaches, GasTwinFormer significantly outperforms Fast-FCNN by $20.46\%$ mIoU and DDRNet by $5.56\%$ mIoU.

GasTwinFormer delivers exceptional inference speed of $114.9$ FPS, enabling real-time processing for practical livestock monitoring applications. Our method runs $1.87\times$ faster than Twins PCPVT-s and $2.23\times$ faster than Twins SVT-s, while also outperforming RepViT ($1.36\times$ faster), UperNet ($1.35\times$ faster), and DeepLabV3 ($1.25\times$ faster). Notably, while SegFormer achieves slightly higher FPS ($119.66$), GasTwinFormer delivers superior accuracy with $2.36\%$ better mIoU. Compared to efficient CNN architectures, our method maintains competitive speed while delivering substantially better accuracy: it runs $2.11\times$ slower than BiSeNetV1 but achieves $21.6\%$ better mIoU.

\noindent \textbf{Dietary Classification Performance.} GasTwinFormer achieves perfect dietary classification accuracy of $100\%$ across all test samples, matching the performance of several state-of-the-art methods including Gasformer, SegFormer, and Twins variants. This demonstrates that our architectural design preserves multi-task learning capability while optimizing segmentation performance. Compared to methods with degraded classification performance, GasTwinFormer outperforms Fast-FCNN by $2.99\%$, BiSeNetV1 by $4.68\%$, and BiSeNetV2 by $1.92\%$, confirming the effectiveness of our Stage 4 feature extraction strategy for capturing dietary-specific emission patterns.

\noindent \textbf{Qualitative Comparison.} ~\Cref{fig:qualitative_comparison} demonstrates distinct performance patterns across methods and dietary treatments. CNN-based BiSeNetV2 consistently produces fragmented predictions with noise artifacts. Transformer methods (GasFormer, SegFormer, Twins PCPVT-S) generate false positives, predicting gas emissions in regions where ground truth shows none, particularly evident in the MD scenarios, while suffering incomplete coverage in HG and HF treatments, with additional over-segmentation in HF cases. GasTwinFormer maintains accurate delineation in MD, comprehensive coverage in both HG and HF cases despite minor over-segmentation in the latter case.

\subsection{Ablation Studies}
\label{sec:ablation}

We evaluate each component of our GasTwinFormer architecture to validate design decisions. 
Unless specified, all ablations use 7×7 LSA windows, Mix-FFN, LRASPP decoder with F1+F2+F3 branch features, F4 for classification head input, Cross Entropy loss for both tasks, 128 internal decoder channels for segmentation, 256 hidden channels for classification, and LE attention pattern per stage.

\noindent \textbf{Decoder head architecture comparison.} We evaluate five different decoder heads to determine the best architecture for methane segmentation. Each decoder is configured with its standard design parameters reported in literature while maintaining consistent backbone features. As shown in ~\cref{tab:foundation_ablation}, the LR-ASPP decoder achieves the best performance at 73.65\% mIoU while maintaining optimal efficiency. Although ISA~\cite{huang2019interlaced} and ANN~\cite{zhu2019asymmetric} heads require significantly more parameters, they deliver inferior performance, validating our lightweight design approach. 

\noindent \textbf{Mix-FFN vs. Regular FFN.} We compare Mix-FFN against standard feed-forward networks within LSA blocks. ~\Cref{tab:foundation_ablation} demonstrates that Mix-FFN substantially outperforms regular FFN, achieving a 1.58 percentage point improvement. Mix-FFN incorporates spatial inductive bias through $3 \times 3$ depth-wise convolutions, crucial for capturing local spatial relationships without explicit positional encodings.

\noindent \textbf{LR-ASPP Multi-scale feature fusion evaluation.} We systematically evaluate different encoder feature combinations for the branch pathway of our Adaptive LR-ASPP decoder. We test all possible combinations of F1, F2, and F3 features to determine the optimal multi-scale fusion strategy. ~\Cref{tab:foundation_ablation} compares performance across all branch combinations. F1+F2+F3 fusion delivers optimal performance, validating our design choice to utilize features from the first three encoder outputs for multi-scale branch processing. Notably, F2+F3 provides competitive performance with reduced computational cost, while individual feature configurations consistently underperform.

\noindent \textbf{LR-ASPP decoder channel analysis.} We examine how the decoder's internal channel dimension affects LR-ASPP performance. ~\Cref{tab:refinement_ablation} shows performance, FLOPs, and parameters across different channel configurations. Our experiments demonstrate that 128 decoder channels deliver the best segmentation accuracy while maintaining computational efficiency. Increasing channels beyond 128 decreases performance while dramatically increasing computational overhead. For example, 1024 channels achieves the second-best mIoU but still underperforms 128 channels while requiring $2.4\times$ more parameters and $7.7\times$ more FLOPs. 

\noindent \textbf{Classification feature source evaluation.} We examine which encoder stage provides optimal features for dietary classification. ~\Cref{tab:refinement_ablation} compares performance across different encoder stage selections. Stage 4 features achieve the best segmentation performance and perfect classification accuracy, validating our design choice. In contrast, Stage 2 and Stage 3 show progressively reduced performance despite requiring fewer parameters. 

\noindent \textbf{Hybrid attention pattern evaluation.} We systematically test different combinations of locally-grouped self-attention (L) and efficient multi-head attention (E) to identify the optimal hybrid pattern. ~\Cref{tab:refinement_ablation} compares results across six attention configurations. EL-EL-EL-EL pattern achieves the highest performance, slightly outperforming our initial LE-LE-LE-LE baseline. Pure attention patterns demonstrate inferior performance, particularly all local attention configurations. This validates that hybrid attention design is crucial, where efficient attention captures global context first, followed by local attention refinement.

\begin{table}[tpb]
\centering
\scriptsize
\begin{tabular}{@{}l|cccc@{}}
\toprule
\textbf{Configuration} & \textbf{\makecell{mIoU\\(\%)$\uparrow$}} & \textbf{\makecell{Diet Acc.\\(\%)$\uparrow$}} & \textbf{\makecell{Params\\(M)$\downarrow$}} & \textbf{\makecell{FLOPs\\(G)$\downarrow$}} \\
\midrule
\multicolumn{5}{l}{\textit{Decoder Head Types}} \\
\midrule
All-MLP~\cite{xie2021segformer} & 73.42 & 99.94 & 3.548 & 7.591 \\
FCN Head~\cite{long2015fully} & 71.23 & 99.94 & 3.416 & 7.303 \\
ISA Head~\cite{huang2019interlaced} & 70.02 & \textbf{100.0} & 4.666 & \textbf{3.337} \\
ANN Head~\cite{zhu2019asymmetric} & 70.67 & \textbf{100.0} & 5.599 & 3.695 \\
LR-ASPP~\cite{howard2019searching} & \textbf{73.65} & \textbf{100.0} & \textbf{3.348} & 3.508 \\
\midrule
\multicolumn{5}{l}{\textit{LSA Feed-Forward Network}} \\
\midrule
Regular FFN & 72.07 & \textbf{100.0} & \textbf{3.065} & \textbf{3.471} \\
MixFFN & \textbf{73.65} & \textbf{100.0} & 3.085 & 3.508 \\
\midrule
\multicolumn{5}{l}{\textit{LR-ASPP Multi-Scale Feature Fusion}} \\
\midrule
F1 only & 72.70 & \textbf{100.0} & \textbf{3.256} & 3.323 \\
F1+F2 & 72.39 & \textbf{100.0} & 3.285 & 3.443 \\
F2 only & 72.55 & \textbf{100.0} & 3.261 & 3.407 \\
F2+F3 & 72.78 & \textbf{100.0} & 3.326 & 3.140 \\
F1+F3 & 72.60 & \textbf{100.0} & 3.319 & 3.387 \\
F3 only & 72.34 & \textbf{100.0} & 3.297 & \textbf{3.019} \\
F1+F2+F3 & \textbf{73.65} & \textbf{100.0} & 3.348 & 3.508 \\
\bottomrule
\end{tabular}
\vspace{-5pt}
\caption{Foundation architecture component studies establishing core design choices through systematic optimization of decoder head, LSA feed-forward network, and multi-scale feature fusion.}
\label{tab:foundation_ablation}
\vspace{-15pt}
\end{table}

\begin{table}[b]
\vspace{-10pt}
\centering
\scriptsize
\begin{tabular}{@{}l|cccc@{}}
\toprule
\textbf{Configuration} & \textbf{\makecell{mIoU\\(\%)$\uparrow$}} & \textbf{\makecell{Diet Acc.\\(\%)$\uparrow$}} & \textbf{\makecell{Params\\(M)$\downarrow$}} & \textbf{\makecell{FLOPs\\(G)$\downarrow$}} \\
\midrule
\multicolumn{5}{l}{\textit{LR-ASPP Channel Scaling}} \\
\midrule
128 ch & \textbf{73.65} & \textbf{100.0} & \textbf{3.348} & \textbf{3.508} \\
256 ch & 72.32 & \textbf{100.0} & 3.644 & 4.729 \\
512 ch & 72.37 & \textbf{100.0} & 4.630 & 9.309 \\
768 ch & 72.97 & 99.38 & 6.140 & 16.741 \\
1024 ch & 73.03 & 99.77 & 8.174 & 27.025 \\
2048 ch & 72.72 & \textbf{100.0} & 21.555 & 96.684 \\
\midrule
\multicolumn{5}{l}{\textit{Classification Feature Source}} \\
\midrule
Encoder Stage 4 & \textbf{73.65} & \textbf{100.0} & 3.348 & \textbf{3.508} \\
Encoder Stage 3 & 69.52 & \textbf{100.0} & 3.323 & 3.508 \\
Encoder Stage 2 & 71.91 & \textbf{100.0} & \textbf{3.299} & 3.508 \\

\midrule
\multicolumn{5}{l}{\textit{Hybrid Attention Pattern Analysis}$^{\dagger}$} \\
\midrule
LE-LE-LE-LE & 73.65 & \textbf{100.0} & 3.348 & 3.508 \\
EL-EL-EL-EL & \textbf{73.69} & 98.70 & 3.348 & 3.508 \\
LL-LL-LL-LL & 68.97 & 98.59 & \textbf{3.113} & \textbf{3.214} \\
EE-EE-EE-EE & 73.17 & \textbf{100.0} & 3.582 & 3.802 \\
LL-LL-EE-EE & 70.56 & 99.77 & 3.319 & 3.259 \\
EE-EE-LL-LL & 73.60 & \textbf{100.0} & 3.376 & 3.757 \\
\midrule
\bottomrule
\end{tabular}
\vspace{-5pt}
\caption{Architecture refinement and pattern optimization studies including decoder channel scaling, classification feature source selection, and systematic evaluation of hybrid attention patterns.}
\label{tab:refinement_ablation}
\vspace{-10pt}
\end{table}

\begin{table}[t]
\centering
\scriptsize
\begin{tabular}{@{}l|cccc@{}}
\toprule
\textbf{Configuration} & \textbf{\makecell{mIoU\\(\%)$\uparrow$}} & \textbf{\makecell{Diet Acc.\\(\%)$\uparrow$}} & \textbf{\makecell{Params\\(M)$\downarrow$}} & \textbf{\makecell{FLOPs\\(G)$\downarrow$}} \\
\midrule
\multicolumn{5}{l}{\textit{Loss Function Comparison}} \\
\midrule
Cross Entropy Loss & 73.69 & 98.70 & 3.348 & 3.508 \\
Dice Loss & 72.57 & \textbf{100.0} & 3.348 & 3.508 \\
Focal Loss & 70.33 & 98.65 & 3.348 & 3.508 \\
Gaussian Plume Loss & \textbf{73.97} & 99.44 & 3.348 & 3.508 \\
\midrule
\multicolumn{5}{l}{\textit{LSA Window Size Optimization}} \\
\midrule
7×7 & 73.97 & 99.44 & 3.348 & 3.508 \\
5×5 & \textbf{74.47} & \textbf{100.0} & 3.348 & 3.428 \\
3×3 & 74.35 & \textbf{100.0} & 3.348 & \textbf{3.367} \\

\bottomrule
\end{tabular}
\vspace{-5pt}
\caption{Task-specific loss and parameter optimization studies comparing segmentation losses and LSA window size refinement.}
\label{tab:optimization_ablation}
\vspace{-15pt}
\end{table}

\noindent \textbf{Loss function comparison.} We evaluate the effectiveness of Gaussian Plume Weighted Dice Loss~\cite{zhou13high} against standard segmentation losses including Cross Entropy, Dice, and Focal loss for segmentation, while maintaining Cross Entropy for classification. \Cref{tab:optimization_ablation} shows the results for this comparison. Gaussian Plume loss achieves the highest performance at 73.97\% mIoU, outperforming all traditional loss functions. While Dice loss achieves perfect classification accuracy, its segmentation performance lags behind by 1.4 percentage points. Focal loss demonstrates the poorest performance across both tasks. 
This demonstrates that domain-specific physical modeling particularly benefits the segmentation task by leveraging the inherent characteristics of gas plume dynamics.

\noindent \textbf{LSA window size optimization.} Finally, we analyze the influence of LSA window size using our best configuration with EL-EL-EL-EL pattern and Gaussian Plume loss.~\Cref{tab:optimization_ablation} compares performance and efficiency across different window sizes. Our analysis reveals that $5 \times 5$ windows achieve the highest performance at 74.47\% mIoU, outperforming both $3 \times 3$ and baseline $7 \times 7$ windows. The $5 \times 5$ size provides optimal balance between local receptive field coverage and computational efficiency. Moderate window sizes prove most effective for capturing gas plume local structures.

\vspace{-5pt}
\section{Conclusion}
We presented GasTwinFormer, a hybrid vision transformer for livestock methane emission segmentation and dietary classification. Comprehensive benchmarking on our beef cattle methane dataset demonstrates that GasTwinFormer outperforms all state-of-the-art methods, achieving superior segmentation and dietary classification performance with significantly fewer computational requirements. Extensive ablation studies validate our architectural design choices. 
While our current study focuses on beef cattle, the architecture is directly extensible to other ruminant species in free-range grazing and broader gas detection contexts by fine-tuning window sizes and training on species-specific OGI data.
This work establishes a strong foundation for automated livestock emission monitoring and climate mitigation applications.

\vspace{-5pt}
\section*{Acknowledgement}
This work is supported by the National Institute of Food and Agriculture, U.S. Department of Agriculture, under award number 2022-70001-37404, and by the Office of the Vice Chancellor for Research at Southern Illinois University Carbondale.

{
    \small
    \bibliographystyle{ieeenat_fullname}
    \bibliography{main}
}

\end{document}